\documentclass{article}
\usepackage{spconf,amsmath,graphicx}
\usepackage{enumitem}
\usepackage{listings}
\usepackage{multicol}
\usepackage{booktabs} 
\usepackage{color}
\usepackage{caption}

\title{VISION AS AN INTERLINGUA: LEARNING MULTILINGUAL SEMANTIC EMBEDDINGS OF UNTRANSCRIBED SPEECH}
\name{David Harwath, Galen Chuang, and James Glass}
\address{Computer Science and Artificial Intelligence Laboratory\\
Massachusetts Institute of Technology\\
Cambridge, MA 02139, USA \\}

\begin{document}
%
\maketitle
\begin{abstract}
In this paper, we explore the learning of neural network embeddings for natural images and speech waveforms describing the content of those images. These embeddings are learned directly from the waveforms without the use of linguistic transcriptions or conventional speech recognition technology. While prior work has investigated this setting in the monolingual case using English speech data, this work represents the first effort to apply these techniques to languages beyond English. Using  spoken captions collected in English and Hindi, we show that the same model architecture can be successfully applied to both languages. Further, we demonstrate that training a multilingual model simultaneously on both languages offers improved performance over the monolingual models. Finally, we show that these models are capable of performing semantic cross-lingual speech-to-speech retrieval.
\end{abstract}
\begin{keywords}
Vision and language, unsupervised speech processing, cross-lingual speech retrieval
\end{keywords}
\section{Introduction}
\label{sec:intro}
The majority of humans acquire the ability to communicate through spoken natural language before they even learn to read and write. Many people even learn to speak in multiple languages, simply by growing up in a multilingual environment. While strongly supervised Automatic Speech Recognition (ASR), Machine Translation (MT), and Natural Language Processing (NLP) algorithms are revolutionizing the world, they still rely on expertly curated training data following a rigid annotation scheme. These datasets are costly to create, and are a bottleneck preventing technologies such as ASR and MT from finding application to all 7,000 human languages spoken worldwide \cite{ethnologue}.

Recently, researchers have investigated models of spoken language that can be trained in a weakly-supervised fashion by augmenting the raw speech data with multimodal context \cite{harwath_glass_asru_2015, harwath_nips, harwath_acl_2017, drexler_2017, chrupala_2017, alishahi_2017, kamper_2017}. Rather than learning a mapping between speech audio and text, these models learn associations between the speech audio and visual images. For example, such a model is capable of learning to associate an instance of the spoken word ``bridge'' with images of bridges. What makes these models compelling is the fact that their training data does not need to be annotated or transcribed; simply recording people talking about images is sufficient. Although these models have been trained on English speech, it is reasonable to assume that they would work well on any other spoken language since they do not make use of linguistic annotation or have any language-specific inductive bias. Here, we demonstrate that this is indeed the case by training speech-to-image and image-to-speech retrieval models in both English and Hindi with the same architecture. We then train multilingual models that share a common visual component, with the goal of using the visual domain as an interlingua or ``Rosetta Stone'' that serves to provide the languages with a common grounding. We show that the audio-visual retrieval performance of a multilingual model exceeds that of the monolingual models, suggesting that the shared visual context allows for cross-pollination between the representations. Finally, we use these models to directly perform cross-lingual audio-to-audio retrieval, which we believe could be a promising direction for future work exploring visually grounded speech-to-speech translation without the need for text transcriptions or directly parallel corpora.

\section{Relation to Prior Work}
\label{sec:prior}
\textbf{Unsupervised Speech Processing.} State-of-the-art ASR systems are close to reaching human parity within certain domains \cite{xiong_2016}, but this comes at an enormous resource cost in terms of text transcriptions for acoustic model training, phonetic lexicons, and large text corpora for language model training. These exist for only a small number of languages, and so a growing body of work has focused on processing speech audio with little-to-no supervision or annotation. The difficulty faced in this paradigm is the enormous variability in the speech signal: background and environmental noise, microphones and recording equipment, and speaker characteristics (gender, age, accent, vocal tract shape, mood/emotion, etc.) are all manifested in the acoustic signal, making it difficult to learn invariant representations of linguistic units without strong guidance from labels. The dominant approaches cast the problem as joint segmentation and clustering of the speech signal into linguistic units at various granularities. Segmental Dynamic Time Warping (S-DTW) \cite{park_glass_sdtw, jansen_2010, jansen_2011} attempts to discover repetitions of the same words in a collection of untranscribed acoustic data by finding repeated regions of high acoustic similarity. Other approaches use Bayesian generative models at multiple levels of linguistic abstraction \cite{lee_glass_2012, but_2016, kamper_2016}. Neural network models have also been used to learn acoustic feature representations which are more robust to undesirable variation \cite{zhang_2012, renshaw_2015, kamper_2015, thiolliere_2015}.

\textbf{Vision and Language.} Modeling correspondences between vision and language is a rapidly growing field at the intersection of computer vision, natural language processing, and speech processing. Most existing work has focused on still-frame images paired with text. Some have studied correspondence matching between categorical abstractions, such as words and objects \cite{barnard_2003, socher_2010, matuszek_2012, frome_2013, lin_2014, kong_2014}. Recently, interest in caption generation has grown, popularized by \cite{karpathy_2015, vinyals_2015, fang_2015}. New problems within the intersection of language and vision continue to be introduced, such as object discovery via multimodal dialog \cite{guess_what}, visual question answering \cite{antol_2015}, and text-to-image generation \cite{reed_2016}. Other work has studied joint representation learning for images and speech audio in the absence of text data. The first major effort in this vein was \cite{roy_2003}, but until recently little progress was made in this direction as the text-and-vision approaches have remained dominant. In \cite{harwath_nips}, embedding models for images and audio captions were shown to be capable of performing semantic retrieval tasks, and more recent works have studied word and object discovery \cite{harwath_acl_2017} and keyword spotting \cite{kamper_2017}. Other work has analyzed these models, and provided evidence that linguistic abstractions such as phones and words emerge in their internal representations \cite{harwath_acl_2017, drexler_2017, chrupala_2017, alishahi_2017}.

\textbf{Machine Translation.} Automatically translating text from one language into another is a well-established problem. At first dominated by statistical methods combining count-based translation and language models \cite{koehn_2003}, the current paradigm relies upon deep neural network models \cite{bahdanau_2015}. New ideas continue to be introduced, including models which take advantage of shared visual context \cite{specia_2016}, but the majority of MT research has focused on the text-to-text case. Recent work has moved beyond that paradigm by implementing translation between speech audio in the source language and written text in the target language \cite{weiss_2017, duong_2016, bansal_2017}. However, it still relies upon expert-crafted transcriptions, and would still require a text-to-speech post-processing module for speech-to-speech translation.

\section{Models}
\label{sec:models}
\begin{figure}
    \includegraphics[width=1\linewidth]{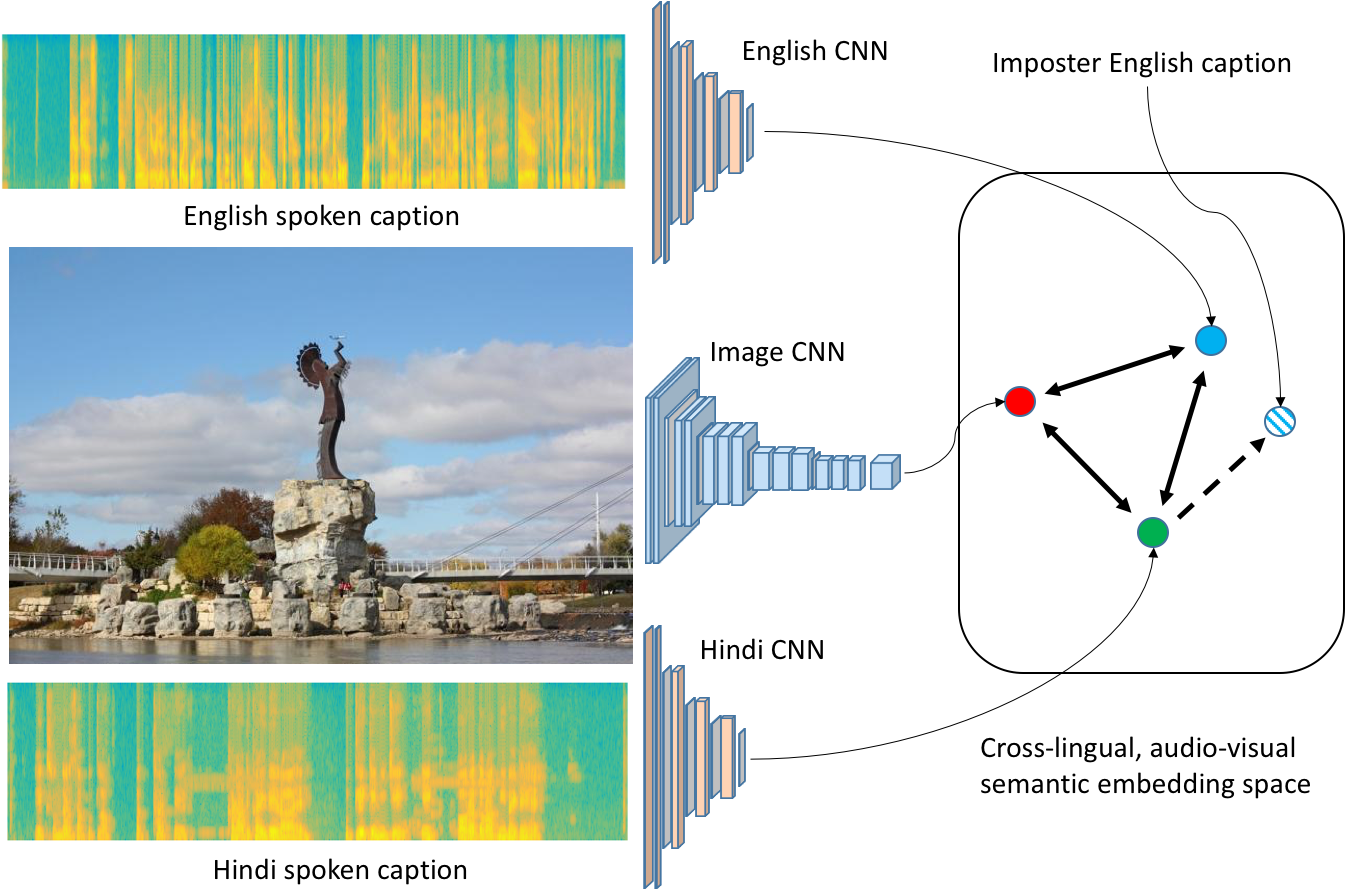}
\caption{Depiction of models and training scheme. Only one impostor point is shown for visual clarity.}
    \label{fig:crosslingual_triple}
\end{figure}

We assume that our data takes the form of a collection of $N$ triples, $(I_i, A^E_i, A^H_i),$ where $I_i$ is the $i^{th}$ image, $A^E_i$ is the acoustic waveform of the English caption describing the image, and $A^H_i$ is the acoustic waveform of the Hindi caption describing the same image. We consider a mapping $F(I_i, A^E_i, A^H_j) \mapsto (e^I_i, e^E_i, e^H_i)$ where $e^I_i, e^E_i, e^H_i \in \mathcal{R}^d$; in other words, a mapping of the image and acoustic captions to vectors in a high-dimensional space. Within this space, our hope is that visual-linguistic semantics are manifested as arithmetic relationships between vectors. We implement this mapping with a set of convolutional neural networks (CNNs) similar in architecture to those presented in \cite{harwath_nips, harwath_acl_2017}. We utilize three networks: one for the image, one for the English caption, and one for the Hindi caption (Figure \ref{fig:crosslingual_triple}).

The image network is formed by taking all layers up to \texttt{conv5} from the pre-trained VGG16 network \cite{vgg}. For a 224x224 pixel, RGB input image, the output of the network at this point would be a downsampled image of width 14 and height 14, with 512 feature channels. We need a means of transforming this tensor into a vector $e^I$ of dimension $d$ (2048 in our experiments), so we apply a linear 3x3 convolution with $d$ filters, followed by global meanpooling. Our input images first resized so that its smallest dimension is 256 pixels, then a random 224x224 pixel crop is chosen (for training; at test time, the center crop is taken), and finally each pixel is mean and variance normalized according to the off-the-shelf VGG mean and variance computed over ImageNet \cite{imagenet}.

The audio architecture we use the same as the one presented in \cite{harwath_acl_2017}, but with the addition of a BatchNorm \cite{batchnorm} layer at the very front of the network, enabling us to do away with any data-space mean and variance normalization; our inputs are simply raw log mel filterbank energies. Our data pre-processing follows \cite{harwath_acl_2017}, where each waveform is represented by a series of 25ms frames with a 10ms shift, and each frame is represented by a vector of 40 log mel filterbank energies. For the sake of enforcing a uniform tensor size within minibatches, the resulting spectrograms are finally truncated or zero-padded to 1024 frames (approx. 10 seconds).

\section{Experiments}
\label{sec:experiments}

\subsection{Experimental Data}
We make use of the Places 205 \cite{places} English audio caption dataset, the collection of which was described in detail in \cite{harwath_nips}, as well as a similar dataset of Hindi audio caption data that we collect via Amazon Mechanical Turk. We use a subset of 85,480 images from the Places data for which we have both an English caption and a Hindi caption. We divide this dataset into a training set of 84,480 image/caption triplets, and a validation set of 1,000 triplets. All captions are completely free-form and unprompted. Turkers are simply shown an image from the Places data and asked to describe the salient objects in several sentences. The English captions on average contain approx. 19.3 words and have an average duration of approx. 9.5 seconds, while the Hindi captions contain an average of 20.4 words and have an average duration of 11.4 seconds.

\subsection{Model Training Procedure}
\begin{table*}[ht!]
\footnotesize
  \caption{Summary of retrieval recall scores for all models. ``English caption'' is abbreviated as E, ``Hindi caption'' as H, and ``Image'' as I. All models were trained with two rounds of 90 epochs, though in all cases they converged before epoch 180. Even though the E$\leftrightarrow$I$\leftrightarrow$H configuration is not specifically trained for the English/Hindi audio-to-audio retrieval tasks, we perform the evaluation anyway for the sake of comparison. Random chance recall scores are R@1=.001, R@5=.005, R@10=.01.}
  \label{tab:retrieval_results}
  \centering
  \setlength{\tabcolsep}{5pt}
  \begin{tabular}{cccccccccccccccccccccc}
    \toprule
    \multicolumn{1}{c}{} & \multicolumn{3}{c}{E $\rightarrow$ I} & \multicolumn{3}{c}{I $\rightarrow$ E} & \multicolumn{3}{c}{H $\rightarrow$ I} & \multicolumn{3}{c}{I $\rightarrow$ H} & \multicolumn{3}{c}{E $\rightarrow$ H} & \multicolumn{3}{c}{H $\rightarrow$ E} \\
    Model & 1 & 5 & 10 & 1 & 5 & 10 & 1 & 5 & 10 & 1 & R5 & 10 & 1 & 5 & 10 & 1 & 5 & 10\\
    \cmidrule(lr){2-4}\cmidrule(lr){5-7}\cmidrule(lr){8-10}\cmidrule(lr){11-13}\cmidrule(lr){14-16}\cmidrule(lr){17-19}
    E$\leftrightarrow$I & .065 & .236 & .367 & .086 & .222 & .343 & - & - & - & - & - & - & - & - & - & - & - & -\\
    H$\leftrightarrow$I & - & - & - & - & - & - & .061 & .185 & .303 & .064 & .186 & .277 & - & - & - & - & - & -\\
    E$\leftrightarrow$H & - & - & - & - & - & - & - & - & - & - & - & - & .011 & .042 & .075 & .013 & .059 & .104\\
    E$\leftrightarrow$I$\leftrightarrow$H & .062 & .248 & .360 & .077 & .247 & .350 & .066 & .205 & .307 & .078 & .208 & .306 & .005 & .012 & .018 & .004 & .016 & .027 \\
    H$\leftrightarrow$E$\leftrightarrow$I$\leftrightarrow$H & .083 & .282 & .424 & .080 & .252 & .365 & .080 & .25 & .356 & .074 & .235 & .354 & .034 & .114 & .182 & .033 & .121 & .203 \\
    \bottomrule
  \end{tabular}
\end{table*}
The objective functions we use to train our models are all based upon a margin ranking criterion \cite{bromley_1994}:
\begin{equation}
    \text{rank}(a, p, i) = \max(0, \eta - s(a, p) + s(a, i))
\end{equation}
where $a$ is the anchor vector, $p$ is a vector ``paired'' with the anchor vector, $i$ is an ``imposter'' vector, $s()$ denotes a similarity function, and $\eta$ is the margin hyperparameter. For a $(a, p, i)$ triplet, the loss is zero when the similarity between $a$ and $p$ is at least $\eta$ greater than the similarity between $a$ and $i$; otherwise, a loss proportional to $s(a, i)$ is incurred. This objective function therefore encourages the anchor and its paired vector to be ``close together,'' and the the anchor to be ``far away'' from the imposter. In all of our experiments, we fix $\eta =1$ and let $s(x,y) = x^Ty$

Given that we have images, English captions, and Hindi captions, we can apply the margin ranking criterion to their neural embedding vectors 6 different ways: each input type can serve as either the anchor point, or as the paired and imposter points. For example, an image embedding may serve as the anchor point, its associated English caption would be the paired point, and an unrelated English caption for some other image would be the imposter point. We can even form composite objective functions by performing multiple kinds of ranking simultaneously. We consider several different training scenarios in Table \ref{tab:retrieval_results}. In each scenario, $\leftrightarrow$ denotes a bidirectional application of the ranking loss function to every tuple within a minibatch of size $B$, e.g. ``English $\leftrightarrow$ Image'' indicates that the terms $\sum_{j=1}^{B}\text{rank}(e^I_j, e^E_j, e^E_k)$ and $\sum_{j=1}^{B}\text{rank}(e^E_j, e^I_j, e^I_l)$ are added to the overall loss, where $k \ne j$ and $l \ne j$ are randomly sampled indices within a minibatch. This is similar to the criteria used in \cite{gella_2017} for multilingual image/text retrieval, except we randomly sample only a single imposter per $(a, p)$ pair.

We trained all models with stochastic gradient descent using a batch size of 128 images with their corresponding captions. All models except the audio-to-audio (no image) were trained with the same learning rate of 0.001, decreased by a factor of 10 every 30 epochs. The audio-to-audio network used an initial learning rate of 0.01, which resulted in instability for the other scenarios. We divided training into two ``rounds'' of 90 epochs (for a total of 180 epochs), where the learning rate is reset back to its initial value starting at epoch 91, and then allowed to decay again. We found this schedule achieved better performance than a single round of 90 epochs, especially for the training scenarios involving simultaneous audio/image and audio/audio retrieval.

\subsection{Audio-Visual and Audio-Audio Retrieval}
For evaluation, we assume a library $L$ of $M$ target vectors, $L = {t_1, t_2, \ldots, t_M}$. Assume we are given a query vector $q$ which is known to be associated with some $t$, but we do not know which; our goal is to retrieve this target from $L$. Given a similarity function $s(q, t)$ (we use $s(q, t) = q^Tt$), we rank all of the target vectors by their similarity to $q$, and retrieve the top scoring 1, 5, and 10.. If the correct target vector is retrieved, a hit is counted; otherwise, we count the result as a miss. With a set of query vectors covering all of $L$ (a set of $M$ vectors containing a $q$ for every $t$), we compute recall scores over the entire query set. Recall that the five training scenarios consider 6 distinct pairwise directions of ranking; for example, we can consider the case in which an English caption is the query and its associated image is the target, or the case in which a Hindi caption is the query and the English caption associated with the same image is the target. We apply the retrieval task to those same directions, and for each model report the relevant recall at the top 1, 5, and 10 results.

Retrieval recall scores for each training scenario are displayed in Table \ref{tab:retrieval_results}.
We found that a small amount of relative weighting was necessary for the H$\leftrightarrow$E$\leftrightarrow$I$\leftrightarrow$H loss function in order to prevent the training from completely favoring audio/image or audio/audio ranking over the other; weighting the E$\leftrightarrow$H ranking loss 5 times higher than that of the E$\leftrightarrow$I and H$\leftrightarrow$I losses produced good results. In all cases, the model trained with the H$\leftrightarrow$E$\leftrightarrow$I$\leftrightarrow$H loss function is the top performer by a significant margin. This suggests that the additional constraint offered by having two separate linguistic accounts of an image's visual semantics can improve the learned representations, even across languages. However, the fact that the E$\leftrightarrow$I$\leftrightarrow$H model offered only marginal improvements over the E$\leftrightarrow$I and H$\leftrightarrow$I models suggests that to take advantage of this additional constraint, it is necessary to enforce semantic similarity between the captions associated with a given image.

Perhaps most interesting are our results on cross-lingual speech-to-speech retrieval. We were surprised to find that the E$\leftrightarrow$H model was able to work at all, given that the retrieval was performed directly on the acoustic level without any linguistic supervision. Even more surprising was the finding that the addition of visual context by the H$\leftrightarrow$E$\leftrightarrow$I$\leftrightarrow$H model approximately doubled the audio-to-audio recall scores across the board, as compared to the E$\leftrightarrow$H model. This suggests that the information contained within the visual modality provides a strong semantic grounding signal that can act as an ``interlingua'' for cross-lingual learning.  To give an example of our model's capabilities, we show the text transcriptions of three randomly selected Hindi captions, along with the transcriptions of their top-1 retrieved English captions using the H$\leftrightarrow$E$\leftrightarrow$I$\leftrightarrow$H model. The English result is denoted by ``E:'', and the approximate translation of the query from Hindi to English is denoted by ``HT:''. Note that the model has no knowledge of any of the ASR text.
\vspace{-2mm}
\begin{figure}[h!]
\footnotesize
\centering
\includegraphics[width=.45\textwidth]{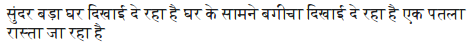}
\captionsetup{labelformat=empty}
\vspace{-3mm}
\caption{HT: ``There is big beautiful house. There is a garden in front of the house. There is a slender road'' \\
E:``A small house with a stone chimney and a porch''
}
\end{figure}
\vspace{-3mm}
\begin{figure}[h!]
\centering
\footnotesize
\includegraphics[width=.45\textwidth]{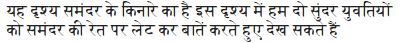}
\captionsetup{labelformat=empty}
\vspace{-3mm}
\caption{HT: ''This is a picture next to the seashore. Two beautiful girls are laying on the sand, talking to each other''\\
E:``A sandy beach and the entrance to the ocean the detail in the sky is very vivid''}
\end{figure}
\vspace{-3mm}
\begin{figure}[h!]
\centering
\footnotesize
\includegraphics[width=.45\textwidth]{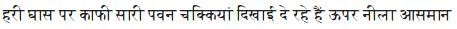}
\captionsetup{labelformat=empty}
\vspace{-3mm}
\caption{HT: ``There are many windmills on the green grass''\\
E:``There is a large windmill in a field''}
\end{figure}
\vspace{-3mm}

\begin{figure}
\vspace{-2mm}
    \includegraphics[width=1\linewidth]{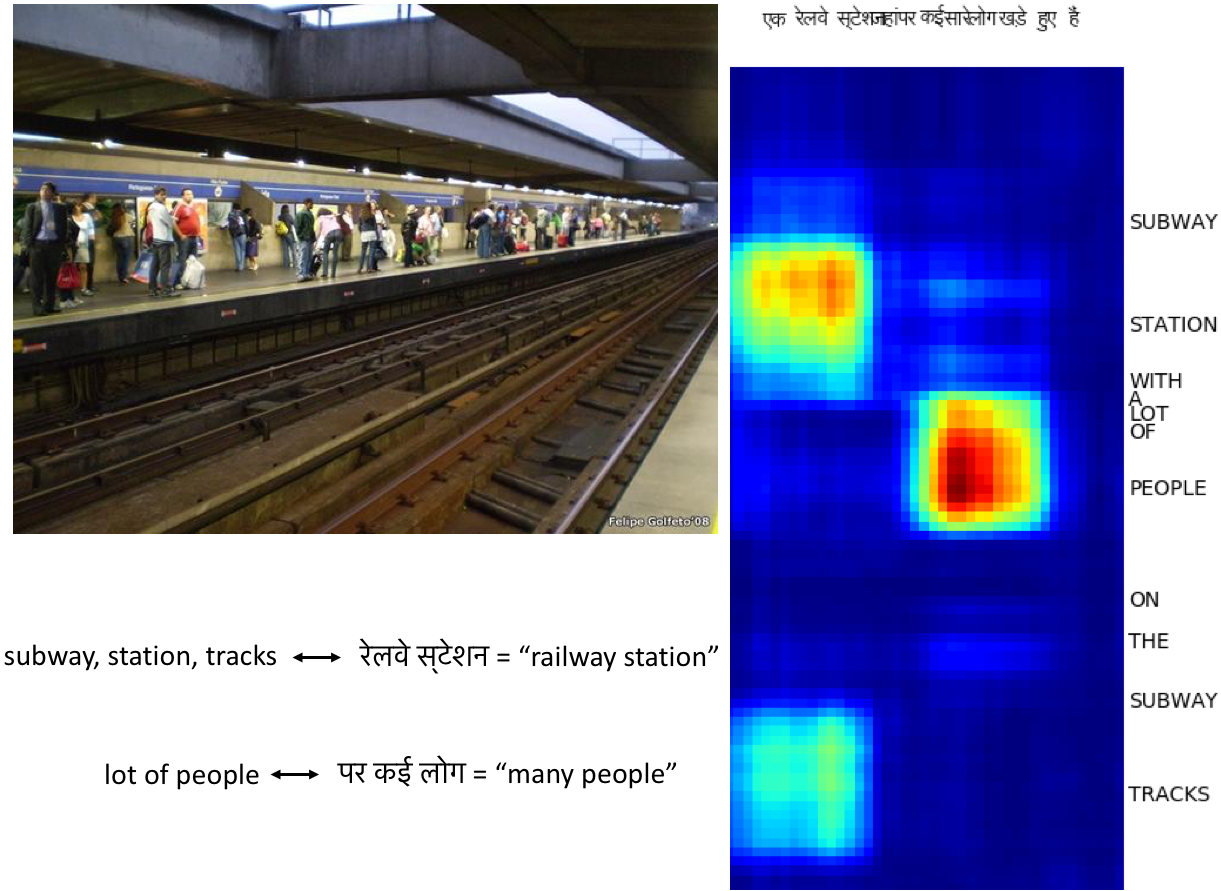}
\caption{Similarity matrix between unpooled embeddings of Hindi and English captions. Transcripts are time-aligned along the axes. Regions of high similarity (red) reflect alignments between the speech signals, which correspond to reasonable translations of the underlying words (bottom left).}
    \label{fig:subway}
\end{figure}

To examine whether individual word translations are indeed being learned, we removed the final pooling layer from the acoustic networks and computed the matrix product of the outputs for the Hindi and English captions associated with the same image. An example of this is shown in Figure \ref{fig:subway}, which seems to indicate that the model is learning approximately word-level translations directly between the speech waveforms. We plan to perform a more objective analysis of this phenomenon in future work.

\section{Conclusion}
\label{sec:conclusion}
In this paper, we applied neural models to associate visual images with raw speech audio in both English and Hindi. These models learned cross-modal, cross-lingual semantics directly at the signal level without any form of ASR or linguistic annotation. We demonstrated that multilingual variants of these models can outperform their monolingual counterparts for speech/image association, and also provided evidence that a shared visual context can dramatically improve the ability of the models to learn cross-lingual semantics. We also provided anecdotal evidence that meaningful word-level translations are being implicitly learned, which we plan to investigate further. We believe that our approach is a promising early step towards speech-to-speech translation models that would not require any form of annotation beyond asking speakers to provide narrations of images, videos, etc. in their native language.

\vfill
\pagebreak

\bibliographystyle{IEEEbib}
\footnotesize
\bibliography{main}

\end{document}